\documentclass{article}
\usepackage{amsmath}
\usepackage{amssymb}
\usepackage{amsfonts}
\usepackage{epsfig}

\begin{document}

\title{Learning ambiguous functions by neural networks}
\author{Rui Ligeiro\thanks{%
INOV INESC -- Instituto de Novas Tecnologias, Rua Alves Redol 9, 1000-029
Lisboa Portugal; rui.ligeiro@inov.pt} \ and R. Vilela Mendes\footnotemark[1] 
\thanks{%
CMAF - Instituto Investiga\c{c}\~{a}o Interdisciplinar, Univ. Lisboa,
Av.Gama Pinto 2, 1649-003 Lisboa, Portugal; vilela@cii.fc.ul.pt,
rvilela.mendes@gmail.com}}
\date{ }
\maketitle

\begin{abstract}
It is not, in general, possible to have access to all variables that
determine the behavior of a system. Having identified a number of variables
whose values can be accessed, there may still be hidden variables which
influence the dynamics of the system. The result is model ambiguity in the
sense that, for the same (or very similar) input values, different objective
outputs should have been obtained. In addition, the degree of ambiguity may
vary widely across the whole range of input values. Thus, to evaluate the
accuracy of a model it is of utmost importance to create a method to obtain
the degree of reliability of each output result. In this paper we present
such a scheme composed of two coupled artificial neural networks: the first
one being responsible for outputting the predicted value, whereas the other
evaluates the reliability of the output, which is learned from the error
values of the first one. As an illustration, the scheme is applied to a
model for tracking slopes in a straw chamber and to a credit scoring model.
\end{abstract}

\section{Introduction}

When dealing with real-world problems, some degree of uncertainty can rarely
be avoided. When modelling physical or social systems, either as a way for
further understanding or as a guide for decision processes, dealing with
uncertainty is a critical issue.

Uncertainty has been formalized in different ways leading to several
uncertainty theories \cite{Klir}. Here we will be concerned with uncertainty
in the construction of models from observed data. In this context
uncertainty may arise either from imprecision in the measurement of the
observed variables or from the fact that the variables that can be measured
do not provide a complete specification of the behavior of the system.

In the context of construction of models of physical phenomena by neural
networks, the problem of learning from data with error bars has been
addressed before by several authors (see for example \cite{Clark} \cite%
{Gabrys}). Here we will be mostly concerned not with imprecision in the
input data but with the fact that the observed variables do not completely
specify the output. In practice this situation is rather complex mainly
because, in general, the uncertainty is not uniform throughout the parameter
space. That is, there might be regions of the parameter space where the
input variables provide an unambiguous answer and others where they are not
sufficient to provide a precise answer. For example in credit scoring, which
we will use here as an example, the "no income, no job, no asset"\footnote{%
Nevertheless many of these so called NINJA scores were financed during the
subprime days} situation is a clear sign of no credit reliability, but most
other situations are not so clear-cut. Therefore it would be desirable to
develop a method that, for each region of parameter space, provides the most
probable outcome but at the same time tells us how reliable the result is.

The purpose of this paper is to develop such a system based on neural
networks that learn in a supervised way. In short, the system consists of
two coupled networks, one to learn the most probable output value for each
set of inputs and the other to provide the expected error (or variance) of
the result for that particular input. The system is formalized as the
problem of learning random functions in the next section. Then we study two
application examples, the first being the measurement of track angles by
straw chambers in high-energy physics and the other a credit scoring model.

\section{Learning the average and variance of random functions}

The general setting which will be analyzed is the following:

The signal to be learned is a random function $y\left( \overset{%
\longrightarrow }{x}\right) $ with distribution $F_{\overset{\longrightarrow 
}{x}}\left( y\right) $. For simplicity we take $y$ to be a scalar and the
index set $\left\{ \overset{\longrightarrow }{x}\right\} $ to be
vector-valued, $\overset{\longrightarrow }{x}\in \mathbb{R}^{i}$. Notice
that we allow for different distribution functions at different points of
the index set.

In the straw chamber example, to be discussed later, $\overset{%
\longrightarrow }{x}$ would be the set of delay times and $y$ the track
angle. For the credit score example, $\overset{\longrightarrow }{x}$ would
be the set of client parameters and $y$ the credit reliability.

In our learning system the $\overset{\longrightarrow }{x}$ values are inputs
to a (feedforward) network with connection strengths $\left\{ W\right\} $,
the output being $f_{W}\left( \overset{\longrightarrow }{x}\right) $. The
aim is to chose a set of connection strengths $\left\{ W\right\} $ that
annihilates the expectation value%
\begin{equation*}
\mathbb{E}\left\{ \sum_{\left\{ \overset{\longrightarrow }{x}\right\}
}\left( f_{W}\left( \overset{\longrightarrow }{x}\right) -y\left( \overset{%
\longrightarrow }{x}\right) \right) ^{2}\right\} =0
\end{equation*}%
However, what, for example, the backpropagation algorithm does is to
minimize $\mathbb{E}\left( f_{W}\left( \overset{\longrightarrow }{x}\right)
-y\left( \overset{\longrightarrow }{x}\right) \right) ^{2}$ for each
realization of the random variable $\overset{\longrightarrow }{x}$. Let us
fix $\overset{\longrightarrow }{x}$ and consider $f_{W}\left( \overset{%
\longrightarrow }{x}\right) $ evolving in learning time. That is, we are
considering, in the learning process, the subprocess corresponding to the
sampling of a particular fixed region of the index set. Then%
\begin{eqnarray*}
f_{W}\left( \overset{\longrightarrow }{x},t+1\right) &=&f_{W}\left( \overset{%
\longrightarrow }{x},t\right) +\frac{\partial f_{W}}{\partial W}\bullet
\Delta W \\
&=&f_{W}\left( \overset{\longrightarrow }{x},t\right) -\eta \frac{\partial
f_{W}}{\partial W}\bullet \frac{\partial e}{\partial W} \\
&=&f_{W}\left( \overset{\longrightarrow }{x},t\right) -2\eta \frac{\partial f%
}{\partial W}\bullet \left( f_{W}\left( \overset{\longrightarrow }{x}\right)
-y\left( \overset{\longrightarrow }{x}\right) \right) \frac{\partial f_{W}}{%
\partial W}
\end{eqnarray*}%
where $\Delta W=-\eta \frac{\partial e}{\partial W}$, $\eta $ being the
learning rate and $e=\left( f_{W}\left( \overset{\longrightarrow }{x}\right)
-y\left( \overset{\longrightarrow }{x}\right) \right) ^{2}$ the error
function.

If the learning rate $\eta $ is sufficiently small for the learning time
scale to be much smaller than the sampling rate of the $y\left( \overset{%
\longrightarrow }{x}\right) $ random variable, the last equality may be
approximated by%
\begin{equation*}
f_{W}\left( \overset{\longrightarrow }{x},t+1\right) =f_{W}\left( \overset{%
\longrightarrow }{x},t\right) -2\eta \frac{\partial f}{\partial W}\bullet
\left( f_{W}\left( \overset{\longrightarrow }{x}\right) -\overline{{y}}%
\left( \overset{\longrightarrow }{x}\right) \right) \frac{\partial f_{W}}{%
\partial W}
\end{equation*}%
$\overline{{y}}\left( \overset{\longrightarrow }{x}\right) $\ denoting the
average value of the random variable $y$ at the argument $\overset{%
\longrightarrow }{x}$. Then a fixed point is obtained for%
\begin{equation*}
f_{W}\left( \overset{\longrightarrow }{x}\right) =\overset{-}{y}\left( 
\overset{\longrightarrow }{x}\right)
\end{equation*}%
That is, $f_{W}\left( \overset{\longrightarrow }{x}\right) $ tends to the
average value of the random variable $y$ at $\overset{\longrightarrow }{x}$.

Similarly if a second network (with output $g_{W^{\prime }}\left( \overset{%
\longrightarrow }{x}\right) $) and the same input $\overset{\longrightarrow }%
{x}$ is constructed according to the learning law%
\begin{equation*}
g_{W^{\prime }}\left( \overset{\longrightarrow }{x},t+1\right) =g_{W^{\prime
}}\left( \overset{\longrightarrow }{x},t\right) +\frac{\partial g_{W^{\prime
}}}{\partial W^{\prime }}\bullet \Delta W^{\prime }
\end{equation*}%
with error function%
\begin{equation*}
e^{\prime }=\left( g_{W^{\prime }}\left( \overset{\longrightarrow }{x}%
\right) -\left( f_{W}\left( \overset{\longrightarrow }{x}\right) -y\left( 
\overset{\longrightarrow }{x}\right) \right) ^{2}\right) ^{2}
\end{equation*}%
and $\Delta W^{\prime }=-\eta ^{\prime }\frac{\partial e^{\prime }}{\partial
W^{\prime }}$, then%
\begin{equation*}
g_{W^{\prime }}\left( \overset{\longrightarrow }{x},t+1\right) =g_{W^{\prime
}}\left( \overset{\longrightarrow }{x},t\right) -2\eta ^{\prime }\frac{%
\partial g_{W^{\prime }}}{\partial W^{\prime }}\bullet \frac{\partial
g_{W^{\prime }}}{\partial W^{\prime }}\left( g_{W^{\prime }}\left( \overset{%
\longrightarrow }{x}\right) -\left( f_{W}\left( \overset{\longrightarrow }{x}%
\right) -y\left( \overset{\longrightarrow }{x}\right) \right) ^{2}\right)
\end{equation*}%
and, under the same assumptions as before concerning the smallness of
learning rates, $g_{W^{\prime }}\left( \overset{\longrightarrow }{x}\right) $
has the fixed point%
\begin{equation*}
g_{W^{\prime }}\left( \overset{\longrightarrow }{x}\right) =\overline{{%
\left( y\left( \overset{\longrightarrow }{x}\right) -\overline{{y}}\left( 
\overset{\longrightarrow }{x}\right) \right) ^{2}}}
\end{equation*}

In conclusion: the first network reproduces the average value of the random
function $y$ for each input $\overset{\longrightarrow }{x}$ and the second
one, receiving as data the errors of the first, reproduces the variance of
the function at $\overset{\longrightarrow }{x}$. Instead of the variance,
the second network might as well be programed to learn the expected value of
the absolute error $\mathbb{E}\left\vert y\left( \overset{\longrightarrow }{x%
}\right) -\overset{-}{y}\left( \overset{\longrightarrow }{x}\right)
\right\vert $. Actually, for numerical convenience, we will use this
alternative in the examples of the next section (Fig.\ref{fig:ann}).

In practice the training of the second network should start after the first
one because, before the first one becomes to converge, its errors are not
representative of the fluctuations of the random function. In general it
seems reasonable to have $\eta ^{\prime }\left( t\right) <\eta \left(
t\right) $with $\eta \left( t\right) $decreasing in time towards a small
fixed value $\neq 0$.

\section{Examples:}

\subsection{Measuring track angles by straw chambers}

\begin{figure}[htb]
\begin{center}
\psfig{figure=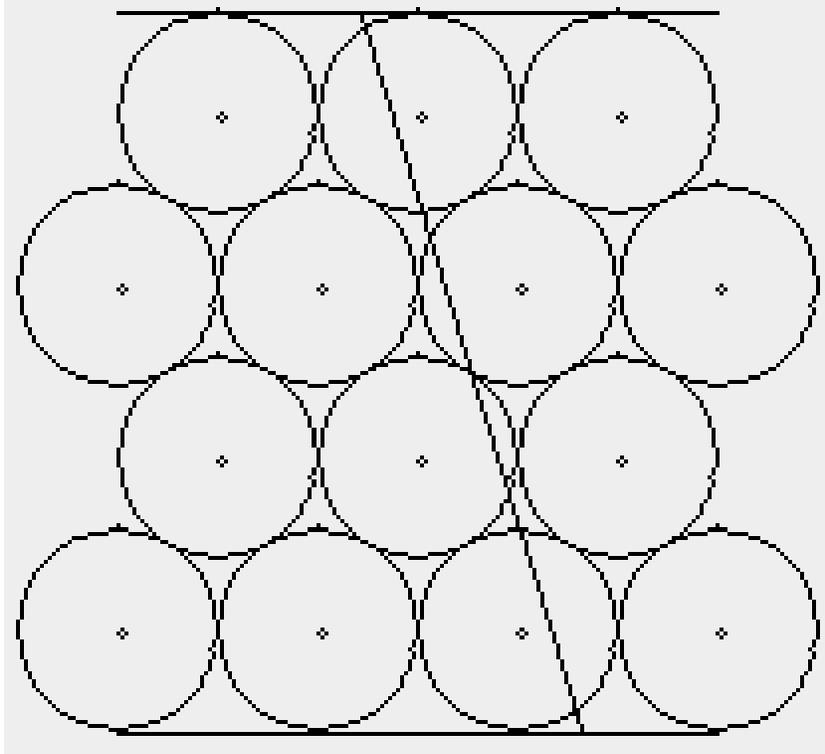,width=11truecm}
\end{center}
\caption{A particle track through a straw chamber. The input values to the neural networks are the delay
times, proportional to the distances of the particle to the wires}
\label{fig:straw_chamber}
\end{figure}

One of the first applications of neural networks to the processing of
high-energy physics data was the work by Denby, Lessner and Lindsey \cite%
{Denby1} on the slopes of particle tracks in straw tube drift chambers. In a
straw chamber (Fig.\ref{fig:straw_chamber}) each wire receives a signal delayed
by a time proportional to the distance of closest approach of the particle
to the wire.

\begin{figure}[htb]
\begin{center}
\psfig{figure=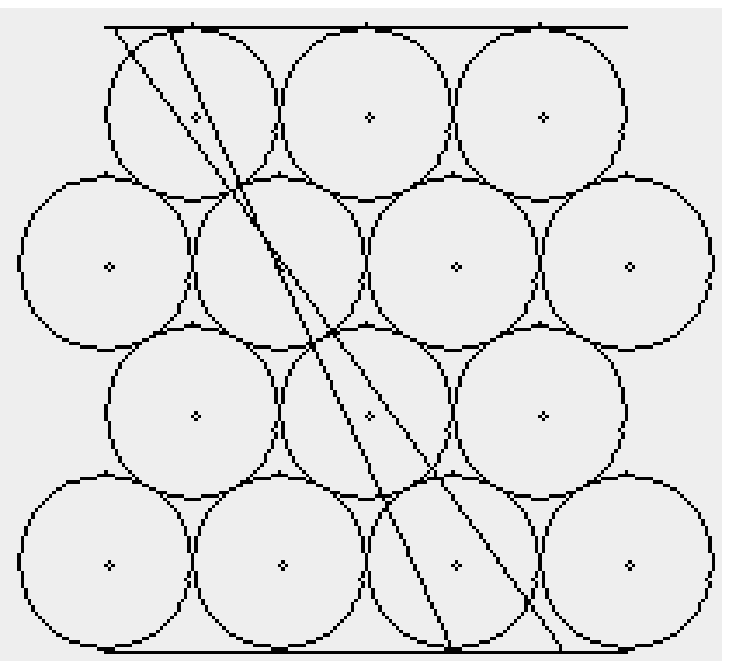,width=11truecm}
\end{center}
\caption{An example with two different beam track angles generating the same
input signal}
\label{fig:ambiguity}
\end{figure}

The neural network receives these times as inputs $\left\{ \overset{%
\longrightarrow }{x}\right\} $, with as many inputs as the number of wires
and, for the training, the track angle $y\left( \overset{\longrightarrow }{x}%
\right) $is the objective function. The half cell shift of alternate layers
in the straw chamber solves some of the left-right ambiguities but this
ambiguity still remains for many directions (Fig.\ref{fig:ambiguity}).

The authors of \cite{Denby1} required the training and test events to pass
through at least four straws to avoid edge effects. Nevertheless they
consistently find large non-Gaussian error tails when testing the trained
network. The authors have not separated the contribution to the tails coming
from the ambiguities from those arising from eventual inadequacies on
training or network architecture. We have repeated the simulations and our
results essentially reproduce those of \cite{Denby1}, showing that the
non-Gaussian tails do indeed originate from the left-right ambiguities. If
edge effects are allowed for, including in the training set events that pass
through less than four straws, the degree of ambiguity and the tails
increase even further.

This example is therefore a typical example of the situation described in
the introduction, where some regions of the input data correspond to a
unique event but others have an ambiguous identification. As the example
shows it is not easy to separate the ambiguous regions from the
non-ambiguous ones because they are mixed all over parameter space. It is
therefore important to have a system that not only provides an answer but
also states how reliable that answer is.

\begin{figure}[htb]
\begin{center}
\psfig{figure=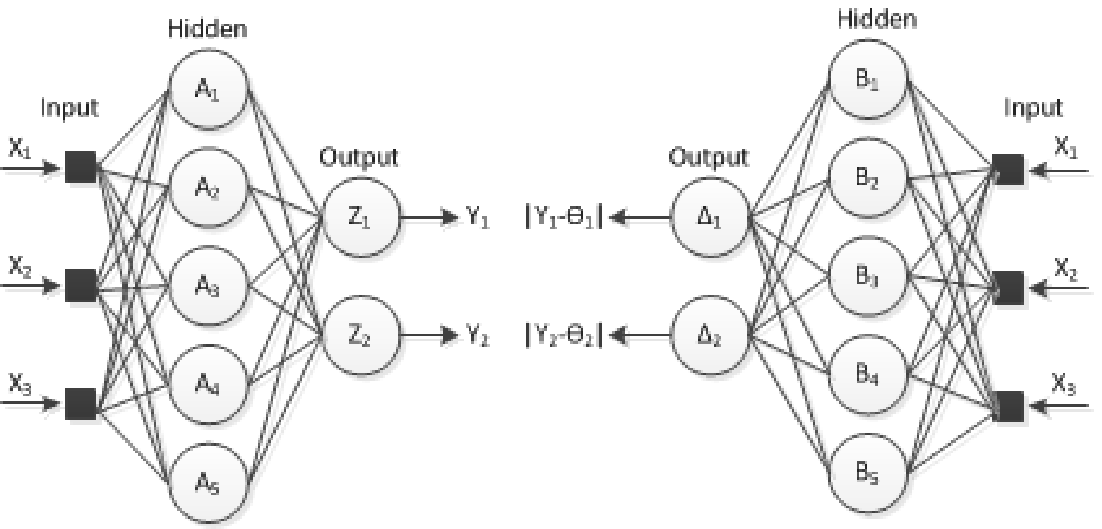,width=11truecm}
\end{center}
\caption{A two-network system that, given
inputs $X_{i}$ and objective values $\protect\theta _{k}$, learns the
average values $\overline{Y_{k}}\left( X_{i}\right) $ and average errors $%
\overline{\left\vert Y_{k}-\theta _{k}\right\vert }$for each set $\left\{
X_{i}\right\} $of input values}
\label{fig:ann}
\end{figure}

We have applied to this example the two-network scheme (Fig.\ref{fig:ann})
described before. Both networks have the same architecture and train using
the same exact input data, the first one (at left in Fig.\ref{fig:ann}) with the
objective track angles and the second (at right in Fig.\ref{fig:ann}) with the
absolute value of the errors of the first. To avoid big fluctuations in
training convergence, the second network starts learning after the first has
stabilized and finished training. Both networks have a feed-forward network
architecture with three neuron layers: input, hidden and output. They both
train using a supervised backpropagation algorithm. The neuron activation
function is the logistic sigmoid.

\begin{figure}[htb]
\begin{center}
\psfig{figure=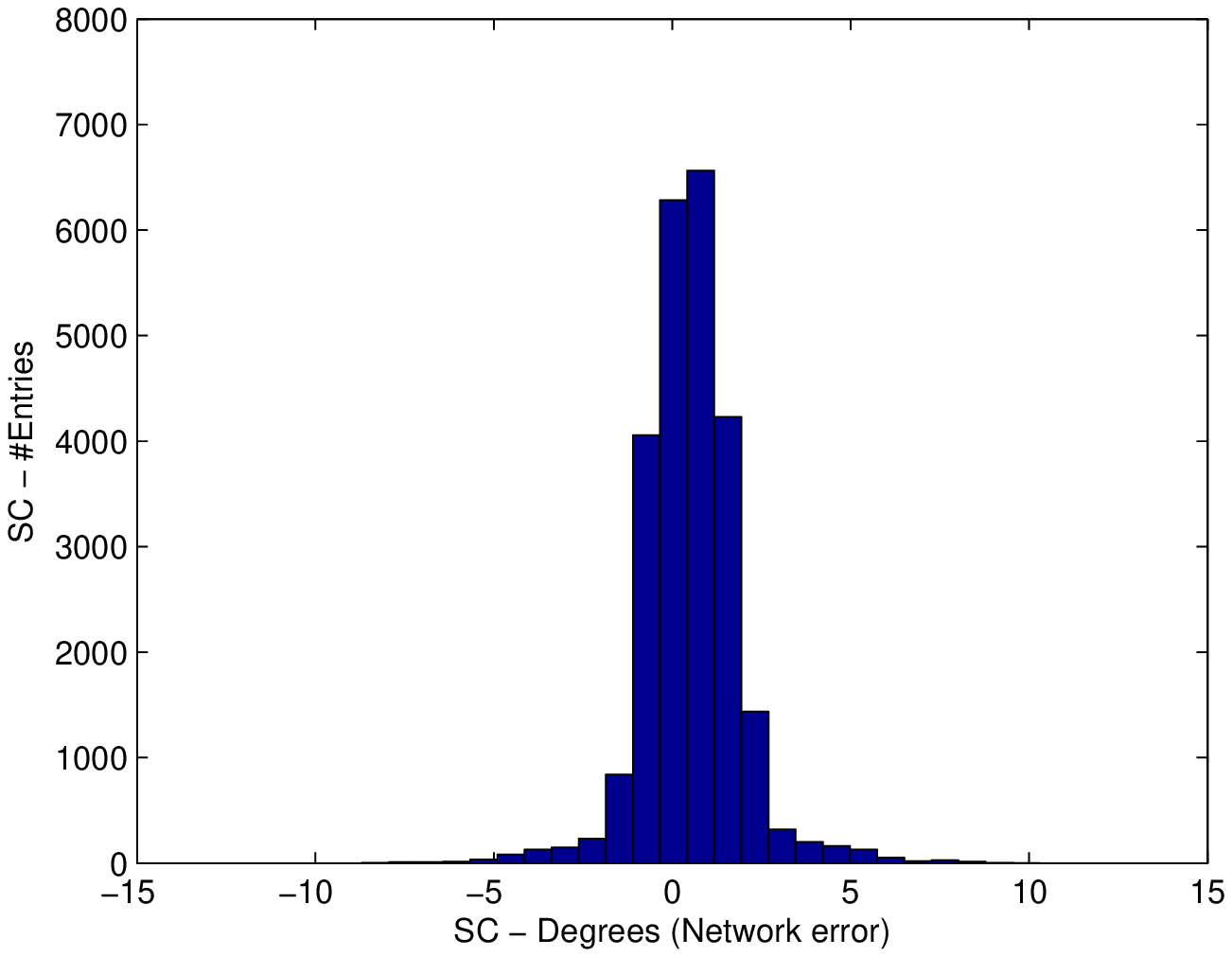,width=11truecm}
\end{center}
\caption{Error distribution in the first network (straw chamber data)}
\label{fig:sc_gauss}
\end{figure}

For the results presented here we use 14 input neurons (representing the
drift times in each straw), 25 hidden neurons and an output neuron for the
slope of each track. We use Monte Carlo generated data coded as follows: If
the track does not meet the straw the input value is zero and if the track
crosses the straw the input value is the difference between the straw radius
and the distance to the wire in the center of the straw. The output is the
angle in degrees of the track slope. A training sample of $25000$ simulated
tracks was generated which trained for $12.5\times 10^{6}$ iterations. After
training, the performance of the network was tested using a new set of $5000$
independent tracks. Fig.\ref{fig:sc_gauss} shows a plot of the first network
errors obtained with the test set. The distribution does present large
non-Gaussian tails because of the left-right ambiguity.

\begin{figure}[htb]
\begin{center}
\psfig{figure=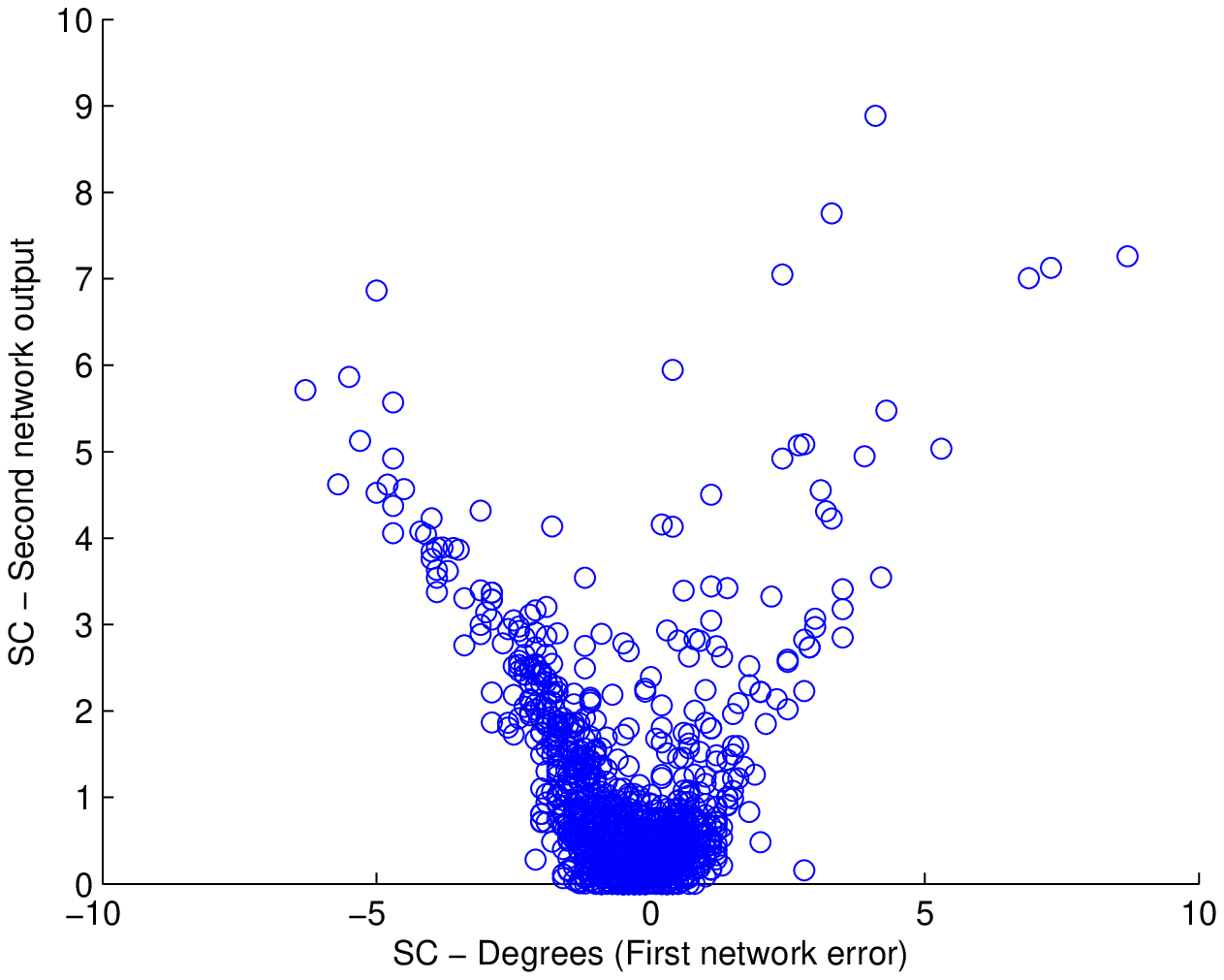,width=11truecm}
\end{center}
\caption{Comparison of the actual error of the first network and the
estimated uncertainty predicted by the second network (straw chamber data)}
\label{fig:sc_error_exp}
\end{figure}

Fig.\ref{fig:sc_error_exp} compares the actual error of the first network with
the uncertainty predicted by the second. One sees that the largest actual
errors do indeed correspond to good uncertainty predictions by the second
network. Of course in a few cases large uncertainty is predicted when the
actual error is small. It only means that particular result is unreliable in
the sense that it was by chance that it fell in the middle of the error bar
interval.

\begin{figure}[htb]
\begin{center}
\psfig{figure=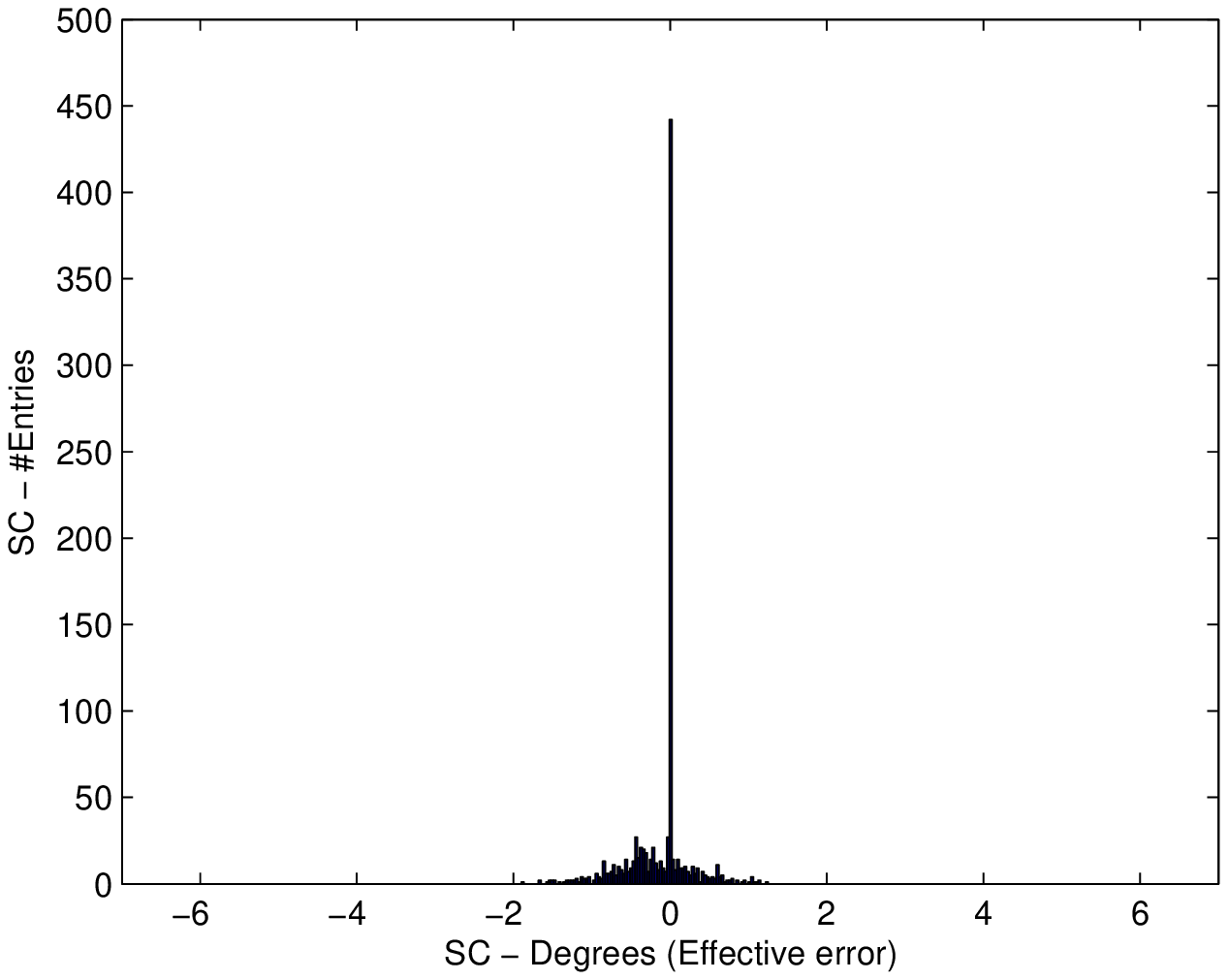,width=11truecm}
\end{center}
\caption{Effective error (straw chamber data)}
\label{fig:sc_prof_delta7}
\end{figure}

Now that we are equipped with a system that predicts both an angle and its
probable uncertainty, it makes sense to state that the result of a
measurement is $\theta \pm \Delta _{ann}$, $\theta $ being the output of the
first network and $\Delta _{ann}$the output of the second. In this sense we
will count an output as an error only when the objective value is outside
the error bars. The \textit{effective error }will be the distance of the
objective value to the boundary of the error bars. Fig.\ref{fig:sc_prof_delta7}
plots the effective error for a sample of $1000$ tracks. Comparison with Fig.%
\ref{fig:sc_gauss} shows how the reliability of the system is improved, because
each time an output value is obtained one has a good estimate of how
accurate it is.

\subsection{A credit scoring model}

Defaulting on loans has recently increased, promoting the search for
accurate techniques of credit evaluation by financial institutions. Credit
scoring is a quantitative method, based on credit report information that
helps lenders in the credit granting decision. The objective is to
categorize credit applicants into two separate classes: the "good credit"\
class, that is, the one likely to repay loans on time and the "bad credit"\
class to which credit should be denied, due to a high probability of
defaulting. For a more detailed understanding of credit scoring models, with
and without neural networks, we refer to \cite{Lando} \cite{Baesens} \cite%
{Lewis}\ \cite{West}.

Here we have developed a credit scoring model based on the two-network
scheme discussed before. Because complete information on the credit
applicants is impossible to obtain and human behavior is dependent on so
many factors, credit scoring is also a typical example of a situation where
one is trying to predict an outcome based on incomplete information.

For the purpose of an open illustration of our system we use here a publicly
available credit data of anonymous clients, downloaded from UCI Irvine
Machine \cite{uci}. It is composed of $1000$ cases, one per applicant, of
which $700$ cases correspond to creditworthy applicants and $300$ cases
correspond to applicants which were later found to be in the bad credit
class. Each instance corresponds to $24$ attributes (e.g., loan amount,
credit history, employment status, personal information, etc.) with the
corresponding credit status of each applicant coded as good ($1$) or bad ($0$%
). Inspecting the database, it is clear that some apparently good attributes
correspond, in the end, to bad credit performance and conversely, putting
into evidence the incomplete information nature of the problem.

\begin{figure}[htb]
\begin{center}
\psfig{figure=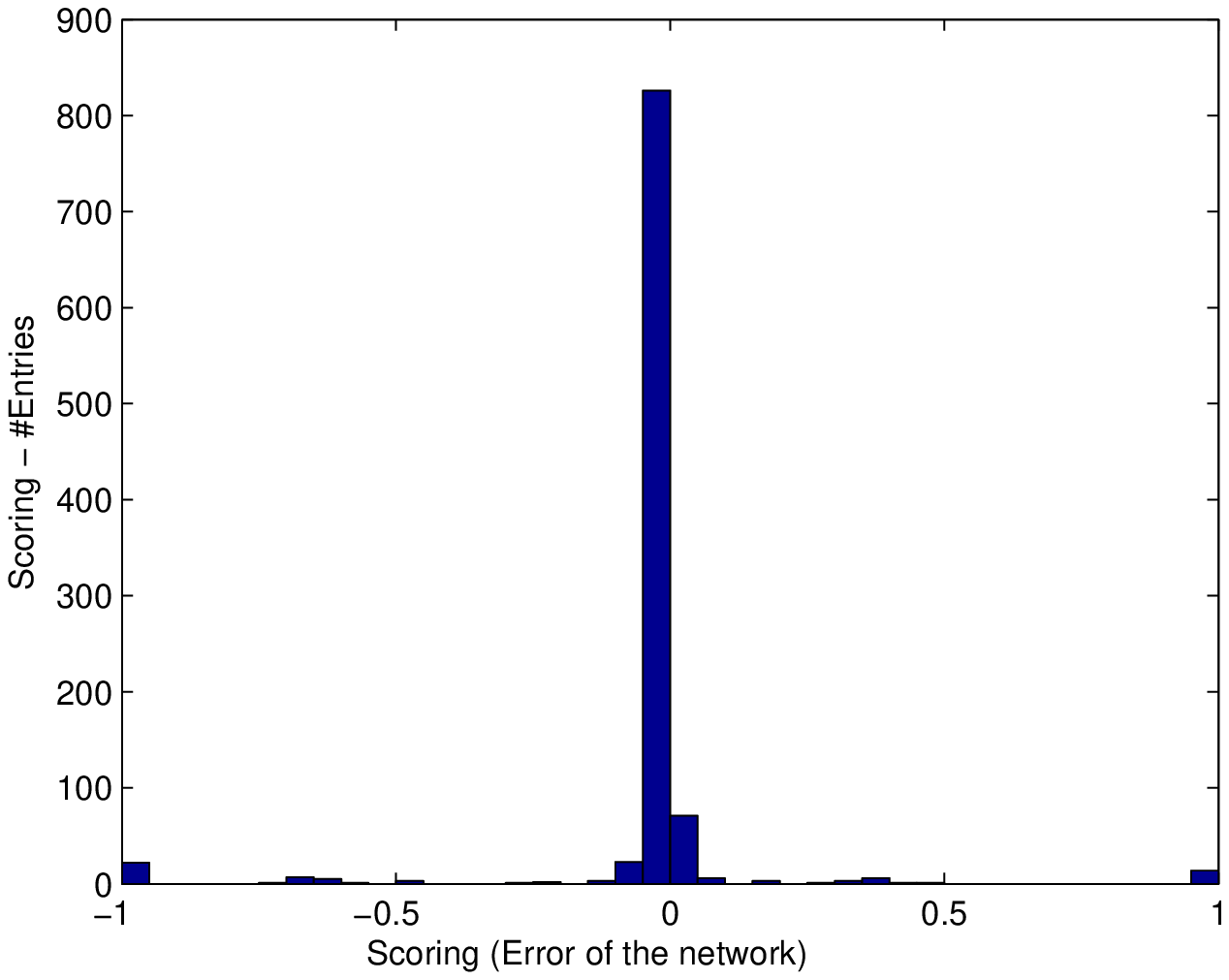,width=11truecm}
\end{center}
\caption{Error distribution in the first network (credit scoring)}
\label{fig:s_gauss}
\end{figure}

For our system the attributes are numerically coded and we use a neural
network architecture with $24$ input neurons (representing the $24$
numerical attributes), $14$ hidden neurons and an output neuron indicating
good or bad credit. The network trained $1000000$ times. To ensure that the
network learns evenly, we randomly alternate between good and bad applicants
instances. After training, the performance of the network was tested. Fig.%
\ref{fig:s_gauss} shows a plot of the errors of the first network after
training. Although, in general, the network provides good estimations, there
are several customers classified as good when they are bad and vice-versa.
In fact, there are some extremely incorrect network predictions, as can
easily been perceived by the bins at the two ends of the histogram. These
bins clearly reveal lack of information in the dataset.

\begin{figure}[htb]
\begin{center}
\psfig{figure=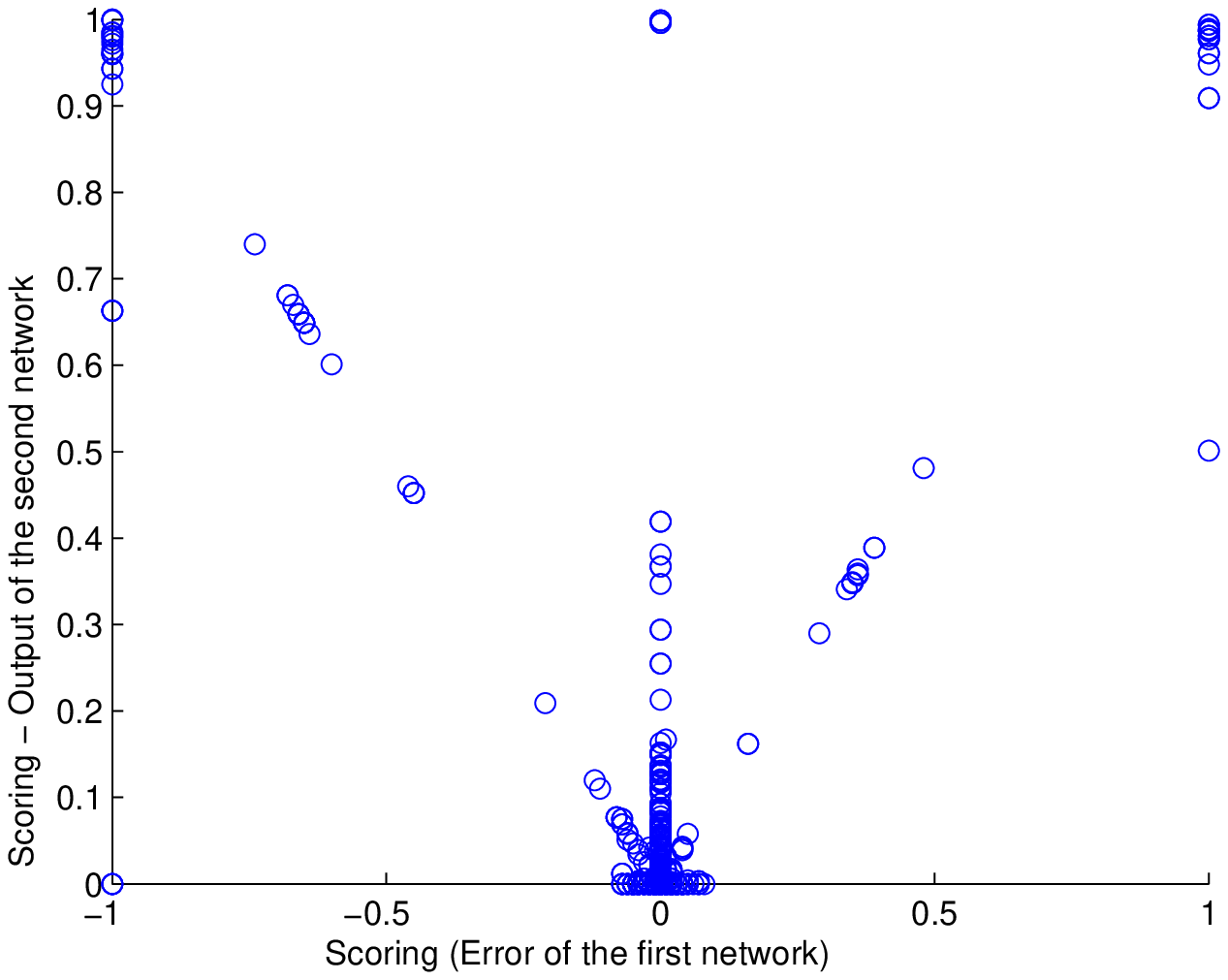,width=11truecm}
\end{center}
\caption{Comparison of the actual error of the first network and the
estimated uncertainty predicted by the second network (credit scoring)}
\label{fig:s_error_exp}
\end{figure}

\begin{figure}[htb]
\begin{center}
\psfig{figure=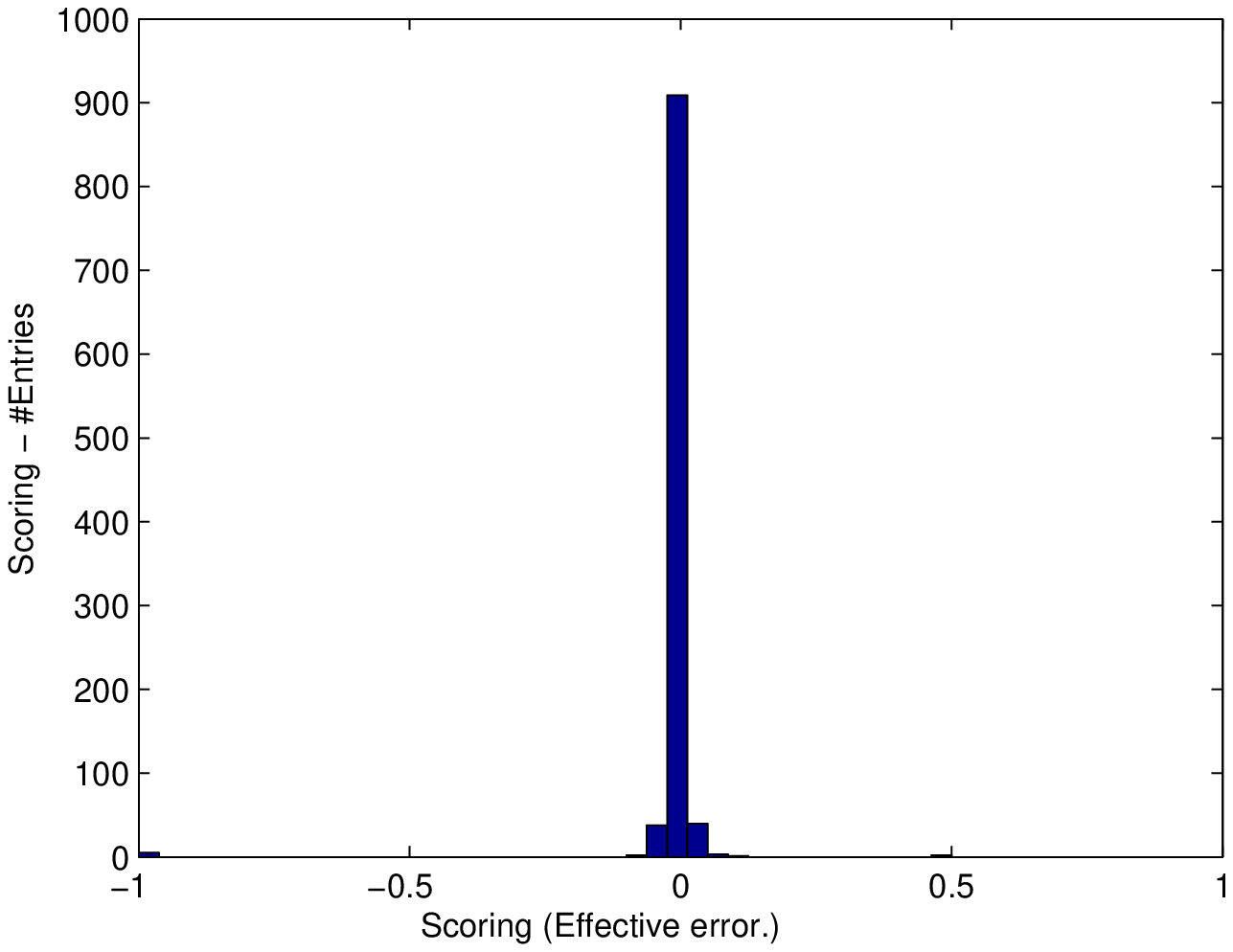,width=11truecm}
\end{center}
\caption{Effective error (credit scoring)}
\label{fig:s_prof_delta}
\end{figure}

As in the previous example, Fig.\ref{fig:s_error_exp} shows the comparison of
the errors in the first network with the estimated uncertainty obtained by
the second network and Fig.\ref{fig:s_prof_delta} shows the effective error
distribution. Similarly to the previous straw chamber example, one obtains
good uncertainty predictions by the second network. The second network
wrongly classified very few cases: only two occurrences with no actual
errors were predicted having maximum uncertainty and only three critical
errors were unsuccessfully predicted without uncertainty.

Looking at the effective error distribution plot, it is easy to confirm the
refinement in the degree of certainty in each estimation. Nevertheless,
there still are a very few occurrences of estimations outside the error bar
interval. Complexity of the human behavior?

\subsection{Conclusions}

The goal of this research was to develop a computational scheme with the
ability to evaluate the degree of reliability of predictive models. Two
application examples were studied, the first one being the measurement of
track angles by straw chambers in high-energy physics and the other a credit
scoring model. Both examples use data with incomplete information. A
two-network system is used which, although not perfect, greatly improves the
reliability check of the predicted results.


\begin{thebibliography}{9}
\bibitem{Klir} G. J. Klir; \textit{Developments in uncertainty-based
information}, Advances in Computers 36 (1993) 255-332.

\bibitem{Clark} K. A. Gernoth and J. W. Clark; \textit{A modified
backpropagation algorithm for training neural networks on data with error
bars}, Computer Physics Commun. 88 (1995) 1-22.

\bibitem{Gabrys} B. Gabrys and A. Bargiela; \textit{Neural network based
decision support in presence of uncertainties}, ASCE\ J. of Water Resources
Planning and Management 125\ (1999) 272-280.

\bibitem{Denby1} B. Denby, E. Lessner and C. S. Lindsey; \textit{Test of
track segment and vertex finding with neural networks}, Proc. 1990 Conf. on
Computing in High Energy Physics, Sante Fe, NM. 1990, AIP Conf Proc. 209,
211.

\bibitem{Lando} D. Lando; \textit{Credit Risk Modeling}, Princeton U. P.
2004.

\bibitem{Baesens} T. Van Gestel and B. Baesens; \textit{Credit Risk
Management: Basic concepts: Financial risk components, Rating analysis,
models, economic and regulatory capital}, Oxford, 2009.

\bibitem{Lewis} E. M. Lewis; \textit{An Introduction to Credit Scoring, }%
Fair, Isaac and Co., San Rafael 1992.

\bibitem{West} D. West; \textit{Neural network credit scoring models},
Computers and Operations Research 27 (2000) 1131-1152.

\bibitem{uci} UCI Machine Learning Repository; URL
http://archive.ics.uci.edu/ml/
\end{thebibliography}
\end{document}